\documentclass{article}
\pdfoutput=1
\PassOptionsToPackage{numbers,compress,square,sort,comma}{natbib}

     \usepackage[preprint]{neurips_2019}

\usepackage{microtype}
\usepackage{graphicx}
\usepackage{booktabs} 
\usepackage{comment}
\usepackage{multirow}
\usepackage[flushleft]{threeparttable}
\usepackage[utf8]{inputenc} 
\usepackage[T1]{fontenc}    
\usepackage{hyperref}       
\usepackage{url}            
\usepackage{breakurl}

\usepackage{amsfonts}       
\usepackage{nicefrac}       
\usepackage{subcaption}
\usepackage{algorithm}
\usepackage{algorithmic}
\usepackage{amssymb}
\usepackage{amsmath}
\usepackage[title]{appendix}
\usepackage{bm}

\newlength\myindent
\title{Pushing the limits of RNN Compression}

\author{%
  Urmish Thakker, Igor Fedorov, Jesse Beu, Dibakar Gope, Chu Zhou \\
  \textbf{Ganesh Dasika\thanks{Currently at AMD Research}  ,Matthew Mattina} \\
  Arm ML Research Lab  
}
\begin{document}

\maketitle

\begin{abstract}
Recurrent Neural Networks (RNN) can be difficult to deploy on resource constrained devices due to their size. As a result, there is a need for compression techniques that can significantly compress RNNs without negatively impacting task accuracy. This paper introduces a method to compress RNNs for resource constrained environments using Kronecker product (KP). KPs can compress RNN layers by $16-38\times$ with minimal accuracy loss. 
We show that KP can beat the task accuracy achieved by other state-of-the-art compression techniques across 4 benchmarks spanning 3 different applications, while simultaneously improving inference run-time. 
\end{abstract}
\section{Introduction}
\label{sec:intro}
Recurrent Neural Networks (RNNs) achieve state-of-the-art (SOTA) accuracy for many applications that use time-series data. As a result, RNNs can benefit important Internet-of-Things (IoT) applications like wake-word detection \cite{zhang2017}, human activity recognition \cite{hammerla2016deep,opp}, and predictive maintenance. IoT applications typically run on highly constrained devices. Due to their energy, power, and cost constraints, IoT devices frequently use low-bandwidth memory technologies and smaller caches compared to desktop and server processors. For example, some IoT devices have 2KB of RAM and 32 KB of Flash Memory. The size of typical RNN layers can prohibit their deployment on IoT devices or reduce execution efficiency \cite{urmtha01RNN}. Thus, there is a need for a compression technique that can drastically compress RNN layers without sacrificing the task accuracy. 

First, we study the efficacy of traditional compression techniques like pruning \cite{suyog} and low-rank matrix factorization (LMF) \cite{DBLP:journals/corr/KuchaievG17,lmf-good1}. We set a compression target of $15\times$ or more and observe that neither pruning nor LMF can achieve the target compression without significant loss in accuracy. We then investigate why traditional techniques fail, focusing on their influence on the rank and condition number of the compressed RNN matrices. We observe that pruning and LMF tend to either decrease matrix rank or lead to ill-condition matrices and matrices with large singular values. 

To remedy the drawbacks of existing compression methods, we propose to use Kronecker Products (KPs) to compress RNN layers. We refer to the resulting models as KPRNNs. We are able to show that our approach achieves SOTA compression on IoT-targeted benchmarks without sacrificing wall clock inference time and accuracy.

\section{Related work}
\label{sec:relwork}

KPs have been used in the deep learning community in the past \cite{kron1,kron2}. For example, \cite{kron2} use KPs to compress fully connected (FC) layers in AlexNet. We deviate from \cite{kron2} by using KPs to compress RNNs and, instead of learning the decomposition for fixed RNN layers, we learn the KP factors directly. 
Additionally, \cite{kron2} does not examine the impact of compression on inference run-time. In \cite{kron1}, KPs are used to stabilize RNN training through a unitary constraint. A detailed discussion of how the present work differs from \cite{kron1} can be found in Section \ref{sec:kprnn}. 

The research in neural network (NN) compression can be roughly categorized into 4 topics: pruning \cite{suyog}, structured matrix based techniques \cite{circular1}, quantization \cite{Quant-hubara,dib1} and tensor decomposition \cite{DBLP:journals/corr/KuchaievG17,hmd}. 
Compression using structured matrices translates into inference speed-up, but only for matrices of size $2048\times 2048$ and larger \cite{NIPS2018_8119} on CPUs or when using specialized hardware \cite{circular1}. As such, we restrict our comparisons to pruning and tensor decomposition. 

\section{Kronecker Product Recurrent Neural Networks}
\label{sec:kprnn}

\subsection{Background}
\label{sec:background}
Let $A \in \mathbb{R}^{m\times n}$, $B \in \mathbb{R}^{m_1 \times n_1}$ and $C \in \mathbb{R}^{m_2\times n_2}$. Then, the KP between $B$ and $C$ is given by
\begin{equation}
    A = B\otimes C
    \label{eq:kp}
\end{equation}
\[
A=
\left[ {\begin{array}{cccc}
b\textsubscript{1,1}\circ C & b\textsubscript{1,2}\circ C & ... & b\textsubscript{1,$n_1$}\circ C\\
b\textsubscript{2,1}\circ C & b\textsubscript{2,2}\circ C & ... & b\textsubscript{2,$n_1$}\circ C\\
. & . & . & . \\
b\textsubscript{$m_1$,1}\circ C & b\textsubscript{1,2}\circ C & ... & b\textsubscript{$m_1$,$n_1$}\circ C\\
\end{array} } \right]
\]
where, $m = m_1 \times m_2$, $n = n_1 \times n_2$, and $\circ$ is the hadamard product. The variables B and C are referred to as the Kronecker factors of A. The number of such Kronecker factors can be 2 or more. If the number of factors is more than 2, we can use  \eqref{eq:kp} recursively to calculate the resultant larger matrix. For example, in the following equation - 
\begin{equation}
    W = W1\otimes W2 \otimes W3
    \label{eq:kp2}
\end{equation}
W can be evaluated by first evaluating $W2 \otimes W3$ to a partial result, say $R$, and then evaluating $W = W1 \otimes R$. 

Expressing a large matrix A as a KP of two or more smaller Kronecker factors can lead to significant compression. For example, $A \in \mathbb{R}^{154 \times 164}$ can be decomposed into Kronecker factors $B \in \mathbb{R}^{11\times 41}$ and $C \in \mathbb{R}^{14\times 4}$.
The result is a $50\times$ reduction in the number of parameters required to store $A$.
Of course, compression can lead to accuracy degradation, which motivates the present work.

\subsection{Prior work on using KP to stabilize RNN training flow}

Jose et al. \cite{kron1} used KP to stabilize the training of vanilla RNN. An RNN layer has two sets of weight matrices - input-hidden and hidden-hidden (also known as recurrent). Jose et al. \cite{kron1} use Kronecker factors of size $2\times 2$ to replace the hidden-hidden matrices of every RNN layer. Thus a traditional RNN cell, represented by:
\begin{gather}
    h_t = f([W_x\; \;W_h]*[x_t; h_{t-1}]) 
    \label{eq:origrnn}
\end{gather}
is replaced by,
\begin{gather}
    h_t = f([W_x\;\;\; W_{0} \otimes W_{1}...\otimes W_{F-1}]*[x_t; h_{t-1}]) \label{eq:origkp} 
\end{gather}
where $W_x$ (input-hidden matrix) $\in \mathbb{R}^{m\times n}$, $W_h$ (hidden-hidden or recurrent matrix) $\in \mathbb{R}^{m\times m}$, $W_{i} \in \mathbb{R}^{2\times 2}$ for $i \in \lbrace 0,\cdots,F-1\rbrace$, $x_{t} \in \mathbb{R}^{n\times 1}$, $h_{t} \in \mathbb{R}^{m\times 1}$, and $F=log_{2}(m)=log_{2}(n)$. Thus a $256\times 256$ sized matrix is expressed as a KP of 8 matrices of size $2\times 2$. For an RNN layer with input and hidden vectors of size 256, this can potentially lead to $\sim 2\times$ compression (as we only compress the $W_h$ matrix). The aim of Jose et al. \cite{kron1} was to stabilize RNN training to avoid vanishing and exploding gradients. They add a unitary constraint to these $2\times 2$ matrices, stabilizing RNN training. However, in order to regain baseline accuracy, they needed to increase the size of the RNN layers significantly, leading to more parameters being accumulated in the $W_x$ matrix in \eqref{eq:origkp}. Thus, while they achieve their objective of stabilizing vanilla RNN training, they achieve only minor compression ($<2\times$).  In this paper, we show how to use KP to compress both the input-hidden and hidden-hidden matrices of vanilla RNN, LSTM and GRU cells and achieve significant compression ($>16\times$). We show how to choose the size and the number of Kronecker factor matrices to ensure high compression rates , minor impact on accuracy, and inference speed-up over baseline on an embedded CPU. 

\subsection{KPRNN Layer}

\paragraph{Choosing the number of Kronecker factors:} A matrix expressed as a KP of multiple Kronecker factors can lead to significant compression. However, deciding the number of factors is not obvious. We started by exploring the framework of \cite{kron1}. We used $2\times 2$ Kronecker factor matrices for hidden-hidden/recurrent matrices of LSTM layers of the key-word spotting network \cite{zhang2017}. This resulted in an approximately $2\times$ reduction in the number of parameters. However, the accuracy dropped by 4\% relative to the baseline. When we examined the $2\times 2$ matrices, we observed that, during training, the values of some of the matrices hardly changed after initialization. This behavior may be explained by the fact that the gradient flowing back into the Kronecker factors vanishes as it gets multiplied with the chain of $2\times2$ matrices during back-propagation. In general, our observations indicated that as the number of Kronecker factors increased, training became harder, leading to significant accuracy loss when compared to baseline.
 
 \begin{algorithm}[htb]
   \caption{Implementation of matrix vector product, when matrix is expressed as a KP of two matrices}
   \label{alg:kpmv}
   \textbf{Input}: Matrices $B$ of dimension $m_1 \times n_1$, $C$ of dimension $m_2 \times n_2$ and $x$ of dimension $n \times 1$. $m = m_1\times m_2$, $n = n_1\times n_2$ \\
   \textbf{Output}: Matrix $y$ of dimension $m\times 1$ 
   \begin{algorithmic}[1]
   \STATE $X = reshape(x,n_2,n_1)$ \COMMENT{reshapes the x vector to a matrix of dimension $n_2\times n_1$}
   \STATE $Bt = B.transpose()$
   \STATE $Y = C\times X\times Bt$
   \STATE $y = reshape(Y,m,1)$ \COMMENT{reshapes the y vector to a matrix of dimension $m\times 1$}
\end{algorithmic}
\end{algorithm}
 
Additionally, using a chain of $2\times 2$ matrices leads to significant slow-down during inference on a CPU. For inference on IoT devices, it is safe to assume that the batch size will be one. When the batch size is one, the RNN cells compute matrix vector products during inference. To calculate the matrix-vector product, we need to multiply and expand all of the $2\times 2$ to calculate the resultant larger matrix, before executing the matrix vector multiplication. Referring to \eqref{eq:origkp}, we need to multiply $W_0,..,W_F$ to create $W_h$ before executing the operation $W_h*h_{t-1}$. The process of expanding the Kronecker factors to a larger matrix, followed by matrix-vector products, leads to a slower inference than the original uncompressed baseline. Thus, inference for RNNs represented using \eqref{eq:origrnn} is faster than the compressed RNN represented using \eqref{eq:origkp}. The same observation is applicable anytime the number of Kronecker factors is greater than $2$.  The slowdown with respect to baseline increases with the number of factors and can be anywhere between $2 - 4\times$.
 
However, if the number of Kronecker factors is restricted to two, we can avoid expanding the Kronecker factors into the larger matrix and achieve speed-up during inference. Algorithm \ref{alg:kpmv} shows how to calculate the matrix vector product when the matrix is expressed as a KP of two Kronecker factors. The derivation of this algorithm can be found in \cite{kpmv}.

\iftrue
\begin{algorithm}[tb]
   \caption{Finding dimension of Kronecker Factors for a matrix of dimension $m\times n$}
   \label{alg:primefactors}
   \textbf{Input}: $list1$ is the sorted list of prime factors of $m$, $list2$ is the sorted list of prime factors of $n$ \\
   \textbf{Output}: $listA$ - Dimension of the first Kronecker factor. $listB$ - Dimension of the second Kronecker factor
   \begin{algorithmic}[1]
   \STATE function reduceList (inputList)
        \STATE \hskip1.0em temp1 = inputList[0]
        \STATE \hskip1.0em      inputList.del(0) //Delete the element at position zero
        \STATE \hskip1.0em      inputList[0] = inputList[0]*temp1
        \STATE \hskip1.0em      inputList.sort('ascending')
        \STATE \hskip1.0em return inputList
   \STATE list2, list1 = reduceList(list2), reduceList(list1).sort('descending')
   \STATE listA, listB = [list1[0],list2[0]], listB = [list1[1],list2[1]]
\end{algorithmic}
\end{algorithm}
\fi
\paragraph{Choosing the dimensions of Kronecker factors:} A matrix can be expressed as a KP of two Kronecker factors of varying sizes. The compression factor is a function of the size of the Kronecker factors. For example, a $256\times 256$ matrix can be expressed as a KP of $2\times 2$ and $128\times 128$ matrices, leading to a $4\times$ reduction in the number of parameters used to store the matrix. However, if we use Kronecker factors of size $32 \times 8$ and $8 \times 32$, we achieve a compression factor of $128$. In this paper, we choose the dimensions of the factors to achieve maximum compression using Algorithm \ref{alg:primefactors}. 

\paragraph{Compressing LSTMs, GRUs and RNNs using the KP:} KPRNN cells are RNN, LSTM and GRU cells with all of the matrices compressed by replacing them with KPs of two smaller matrices. For example, the RNN cell depicted in \eqref{eq:origrnn} is replaced by: 
\begin{gather}
    KPRNN\;\;cell: h_t = f((W_{1} \otimes W_{2})*[x_t; h_{t-1}])
    \label{eq:kprnn}
\end{gather}
where $x_{t} \in \mathbb{R}^{n\times 1}$, $h_{t} \in \mathbb{R}^{m\times 1}$, $W_{1} \in \mathbb{R}^{m_1\times n1}$, $W_{2} \in \mathbb{R}^{m_2\times n2}$, $m_1\times m_2 = m$ and $n_1\times n_2 = (m+n)$. LSTM, GRU and FastRNN cells are compressed in a similar fashion. Instead of starting with a trained network and decomposing its matrices into Kronecker factors, we replace the RNN/LSTM/GRU cells in a NN with its KP equivalent and train the entire model from the beginning.

\section{Results}
\label{sec:results}

\begin{table*}
\centering
\caption{Benchmarks evaluated in this paper. These benchmarks represent some of the key applications in the IoT domain - Image Classification, Key-word spotting, Bidirectional LSTM. We cover a wide variety of applications and RNN cell types.}
\label{tab:benchmarks}
\begin{tabular}{|p{0.18\columnwidth}|p{0.18\columnwidth}|p{0.18\columnwidth}|p{0.18\columnwidth}|p{0.18\columnwidth}|}

\hline
 & MNIST- & USPS-& KWS-&  HAR1- \\ 
 & LSTM & FastRNN & LSTM &  BiLSTM\\ \hline
Reference Paper &  &\cite{msr}  &\cite{zhang2017}  &\cite{hammerla2016deep}  \\ \hline
Cell Type & LSTM & FastRNN & LSTM & Bi-LSTM \\ \hline

Dataset & \cite{mnist} & \cite{usps} & \cite{warden} &\cite{opp} \\ \hline 
\end{tabular}
\vspace{-1.0em}
\end{table*}

\begin{table*}[t]
\vspace{0cm}
\centering
\begin{threeparttable}
\caption{Model accuracy and runtime for our benchmarks before and after compression. The baseline networks are compared to networks with RNN layers in the baseline compressed using KPs, magnitude pruning, LMF, or by scaling the network size (Small Baseline). Each compressed network has fewer RNN parameters than the baseline (size indicated). For each row, best results are indicated in bold. The KP-based networks are consistently the most accurate alternative while still having speed-up over the baseline.}
\label{tab:kprnnnresults}
\begin{tabular}{|l|l|l|l|l|l|l|}
\hline
\multirow{2}{*}{Benchmark Name} &
\multirow{2}{*}{Parameter measured} &
\multicolumn{1}{c|}{} & \multicolumn{4}{c|}{Compression Technique} \\ \cline{3-7}
\multicolumn{1}{|c|}{} & \multicolumn{1}{c|}{} & \multicolumn{1}{c|}{\textit{Baseline}} & \multicolumn{1}{c|}{Small Baseline} & \multicolumn{1}{c|}{\begin{tabular}[c]{@{}c@{}}Magnitude\\ Pruning\end{tabular}} & LMF & KP \\  \hline
\multirow{4}{*}{MNIST-LSTM} & Model Size (KB)\tnote{1} & \textit{44.73} & 4.51 & 4.19 & 4.9 & \textbf{4.05} \\ \cline{2-7} 
\multicolumn{1}{|c|}{} & Accuracy (\%) & \textit{99.40} & 87.50 & 96.49 & 97.40 & \textbf{98.44} \\ \cline{2-7} 
\multicolumn{1}{|c|}{} & Compression factor \tnote{2} & \textit{1}$\times$ & $10\times$ & $16.7\times$ & $13.08\times$ & \textbf{17.6}$\bm{\times}$ \\ \cline{2-7} 
\multicolumn{1}{|c|}{} & Runtime (ms) & \textit{6.3} & 0.7 & \textbf{0.66} & 1.8 & 4.6 \\ \hline
\multirow{4}{*}{HAR1-BiLSTM} & Model Size (KB)\tnote{1} & \textit{1462.3} & 75.9 & 75.55 & 76.39 & \textbf{74.90} \\ \cline{2-7} 
 & Accuracy (\%) & \textit{91.90} & 88.84 & 82.97 & 89.94 & \textbf{91.14} \\ \cline{2-7} 
 & Compression factor \tnote{2} & \textit{1}$\times$ & 19.8$\times$ & 28.6$\times$ & 28.1$\times$ & \textbf{29.7}$\bm{\times}$ \\ \cline{2-7} 
 & Runtime (ms) & \textit{470} & \textbf{29.92} & 98.2 & 64.12 & 157 \\ \hline
\multirow{4}{*}{KWS-LSTM} & Model Size (KB)\tnote{1} & \textit{243.4} & 16.3 & 15.56 & 16.79 & \textbf{15.30} \\ \cline{2-7} 
 & Accuracy (\%) & \textit{92.5} & 89.70 & 84.91 & 89.13 & \textbf{91.2} \\ \cline{2-7} 
 & Compression factor \tnote{2} & \textit{1}$\times$ & 15.8$\times$ & 23.81$\times$ & 21.2$\times$ & \textbf{24.47}$\bm{\times}$ \\ \cline{2-7} 
 & Runtime (ms) & \textit{26.8} & \textbf{2.01} & 5.9 & 4.1 & 17.5 \\ \hline
\multirow{4}{*}{USPS-FastRNN} & Model Size (KB)\tnote{1} & \textit{7.25} & 1.98 & 1.92 & 2.04 & \textbf{1.63} \\ \cline{2-7} 
 & Accuracy (\%) & \textit{93.77} & 91.23 & 88.52 & 89.56 & \textbf{93.20} \\ \cline{2-7} 
 & Compression factor \tnote{2} & \textit{1}$\times$ & 4.4$\times$ & 8.94$\times$ & 8$\times$ & \textbf{16}$\bm{\times}$ \\ \cline{2-7} 
 & Runtime (ms) & \textit{1.17} & 0.4 & \textbf{0.375} & 0.28 & 0.6 \\ \hline
\end{tabular}
\begin{tablenotes}
  \item[1] Model size is calculated assuming 32-bit weights. Further opportunities exist to compress the network via quantization and compressing the fully connected softmax layer.
  \item[2] We measure the amount of compression of the LSTM/FastRNN layer of the network
\end{tablenotes}
\end{threeparttable}
\vspace{-0.5cm}
\end{table*}

\paragraph{Other compression techniques evaluated:} We compare networks compressed using KPRNN with three techniques - pruning, LMF and Small Baseline. 

\paragraph{Training platform, infrastructure, and inference run-time measurement:} \label{sec:trainingPI} We use Tensorflow 1.12 as the training platform and 4 Nvidia RTX 2080 GPUs to train our benchmarks. To measure the inference run-time, we implement the baseline and the compressed cells in C++ using the Eigen library and run them on the Arm Cortex-A73 core of a Hikey 960 development board. 

\paragraph{Dataset and benchmarks:} We evaluate the impact of compression using the techniques discussed in Section \ref{sec:kprnn} on a wide variety of benchmarks.
Table \ref{tab:benchmarks} shows the benchmarks used in this work. 



\subsection{KPRNN networks}
\label{sec:results-kprnn}

Table \ref{tab:kprnnnresults} shows the results of applying the KP compression technique across a wide variety of applications and RNN cells. As mentioned in Section \ref{sec:kprnn}, we target the point of maximum compression using two matrix factors.

\subsection{Possible explanation for the accuracy difference between KPRNN, pruning, and LMF} 
In general, the poor accuracy of LMF can be attributed to significant reduction in the rank of the matrix (generally $<10$). KPs, on the other hand, will create a full rank matrix if the Kronecker factors are fully ranked \cite{laub2005matrix}. 
We observe that, Kronecker factors of all the compressed benchmarks are fully-ranked. A full-rank matrix can also lead to poor accuracy if it is ill-conditioned. However, the condition numbers of the matrices of the best-performing KP compressed networks discussed in this paper are in the range of $1.2$ to $7.3$. To prune a network to the same compression factor as KP, networks need to be pruned to 94\% sparsity or above. Pruning FastRNN cells to the required compression factor leads to an ill-conditioned matrix. This may explain the poor accuracy of sparse FastRNN networks. However, for other pruned networks, the resultant sparse matrices have a condition number less than $20$ and are fully-ranked. Thus, condition number does not explain the loss in accuracy for these benchmarks. To further understand the loss in accuracy of pruned LSTM networks, we looked at the singular values of the resultant sparse matrices in the KWS-LSTM network. Let $y = Ax$. The largest singular value of $A$ upper-bounds $\Vert y \Vert_2$, i.e. the amplification applied by $A$. Thus, a matrix with larger singular value can lead to an output with larger norm \cite{linalgbook}. Since RNNs execute a matrix-vector product followed by a non-linear sigmoid or tanh layer, the output will saturate if the value is large. The matrix in the LSTM layer of the best-performing pruned KWS-LSTM network has its largest singular value in the range of $48$ to $52$ while the baseline KWS-LSTM network learns a LSTM layer matrix with largest singular value of $19$ and the Kronecker product compressed KWS-LSTM network learns LSTM layers with singular values less than $15$. This might explain the especially poor results achieved after pruning this benchmark. Similar observations can be made for the pruned HAR1 network.
\section{Conclusion}
\label{sec:conclusion}
We show how to compress RNN Cells by $15\times$ to $38\times$ using Kronecker products. We call the cells compressed using Kronecker products as KPRNNs. KPRNNs can act as a drop in replacement for most RNN layers and provide the benefit of significant compression with marginal impact on accuracy. None of the other compression techniques (pruning, LMF) match the accuracy of the Kronecker compressed networks. We show that this compression technique works across 5 benchmarks that represent key applications in the IoT domain.

\bibliographystyle{plain}

\newpage
\begin{appendices}
\end{appendices}
\end{document}